\pdfoutput=1

\documentclass[11pt]{article}

\usepackage[]{acl}

\usepackage{times}
\usepackage{latexsym}

\usepackage[T1]{fontenc}

\usepackage[utf8]{inputenc}

\usepackage{microtype}

\usepackage{multirow}
\usepackage{comment}
\usepackage{caption}
\title{E2E Refined Dataset}

\author{Keisuke Toyama \quad Katsuhito Sudoh \quad Satoshi Nakamura \\
  Nara Institute of Science and Technology, Japan \\
  \texttt{\{toyama.keisuke.tb5, sudoh, s-nakamura\}@is.naist.jp} }

\begin{document}
\maketitle
\begin{abstract}
Although the well-known MR-to-text E2E dataset has been used by many researchers, its MR-text pairs include many deletion/insertion/substitution errors.
Since such errors affect the quality of MR-to-text systems, they must be fixed as much as possible.
Therefore, we developed a refined dataset and some python programs that convert the original E2E dataset into a refined dataset.
\end{abstract}

\section{Introduction}\label{sec:introduction}

The E2E dataset \citep{novikova2017e2e}, which was used in the E2E NLG Challenge \citep{dusek.etal2020:csl}, is a very popular dataset for natural language generation (NLG) from meaning representation (MR): MR-to-text.
The dataset consists of a set of pairs of a British English sentence and a corresponding MR with eight attributes (\texttt{name}, \texttt{eatType}, \texttt{food}, \texttt{priceRange}, \texttt{customer rating}, \texttt{area}, \texttt{familyFriendly}, and \texttt{near}) in a restaurant recommendation domain (\autoref{tab:example_of_the_e2e_dataset}).
However, some of the MR-text pairs suffer from the following errors: deletion (an MR is not reflected in the text), insertion (an MR whose value is empty appears in the text with an unintended value), and substitution (an MR value is replaced in the text) (\autoref{tab:example_of_mr_errors_in_the_e2e_dataset}).

\begin{table}[t]
    \centering
    \begin{small}
    \begin{tabular}{l|l|l}
    \hline
    \multirow{8}{*}{MR}&\texttt{name}&The Wrestlers\\
    \cline{2-3}
    &\texttt{eatType}&restaurant\\
    \cline{2-3}
    &\texttt{food}&Italian\\
    \cline{2-3}
    &\texttt{priceRange}&moderate\\
    \cline{2-3}
    &\texttt{customer rating}&(empty)\\
    \cline{2-3}
    &\texttt{area}&city centre\\
    \cline{2-3}
    &\texttt{familyFriendly}&yes\\
    \cline{2-3}
    &\texttt{near}&Raja Indian Cuisine\\
    \hline
    \multirow{4}{*}{Text}&\multicolumn{2}{p{6cm}}{The Wrestlers is a moderately priced restaurant that serves Italian food. It is located in the city centre near Raja Indian Cuisine. Great place to bring your family.}\\
    \hline
    \end{tabular}
    \end{small}
    \caption{Example of the E2E dataset}
    \label{tab:example_of_the_e2e_dataset}
    \vspace{5mm}
    \centering
    \begin{small}
    \begin{tabular}{l|l|l}
    \hline
    \multirow{8}{*}{MR}&\texttt{name}&The Punter\\
    \cline{2-3}
    &\texttt{eatType}&coffee shop\\
    \cline{2-3}
    &\texttt{food}&{\bf English}\\
    \cline{2-3}
    &\texttt{priceRange}&{\it moderate}\\
    \cline{2-3}
    &\texttt{customer rating}&1 out of 5\\
    \cline{2-3}
    &\texttt{area}&(empty)\\
    \cline{2-3}
    &\texttt{familyFriendly}&yes\\
    \cline{2-3}
    &\texttt{near}&Caf\'e Sicilia\\
    \hline
    \multirow{2}{*}{Text}&\multicolumn{2}{p{6cm}}{The Punter is a {\it cheap} family friendly coffee shop located in \underline{City Centre} near Caf\'e Sicilia. 1 out of 5 customer rating.}\\
    \hline
    \end{tabular}
    \end{small}
    \caption{Example of MR errors in the E2E dataset: \texttt{food} is deleted (``{\bf English}''), \texttt{area} is inserted (``\underline{city centre}''), and \texttt{priceRange} is replaced (``{\it moderate}'' → ``{\it cheap}'') in the text.}
    \label{tab:example_of_mr_errors_in_the_e2e_dataset}
\end{table}

To properly control the content of a sentence by MR-to-text, such incorrect data must be removed from the dataset.
Although there have been some updates to the E2E dataset to rectify its errors \cite{dusek-etal-2019-semantic,castro-ferreira-etal-2021-enriching},
the updated datasets still include such deletion/insertion/substitution errors.
The specific number of errors that we identified is shown in \autoref{tab:number_of_mr_labelling_errors}.
In this work, we further refine the E2E dataset by additional manual annotation of the MR values.
We fixed the errors in the MR-text correspondences and removed inappropriate data.

\begin{table*}[t]
    \centering
    \begin{small}
    \begin{tabular}{c|r|r|r|r|r|r|r|r|r}
    \multirow{3}{*}{Error type}&\multicolumn{3}{c|}{\multirow{2}{*}{E2E dataset}}&\multicolumn{3}{c|}{Cleaned dataset}&\multicolumn{3}{c}{Enriched dataset}\\
    &\multicolumn{3}{c|}{}&\multicolumn{3}{c|}{\cite{dusek-etal-2019-semantic}}&\multicolumn{3}{c}{\cite{castro-ferreira-etal-2021-enriching}}\\
    \cline{2-10}
    &\multicolumn{1}{c|}{Training}&\multicolumn{1}{c|}{Validation}&\multicolumn{1}{c|}{Test}&\multicolumn{1}{c|}{Training}&\multicolumn{1}{c|}{Validation}&\multicolumn{1}{c|}{Test}&\multicolumn{1}{c|}{Training}&\multicolumn{1}{c|}{Validation}&\multicolumn{1}{c}{Test}\\
    \hline
    Deletion&10,931&1,096&1,315&23&1&1&1,262&145&89\\
    Insertion&10,028&263&16&4,475&471&745&25,570&2,724&3,082\\
    Substitution&9,290&794&945&5,795&616&666&4,172&413&395\\
    \end{tabular}
    \end{small}
    \caption{Number of MR labelling errors in each dataset}
    \label{tab:number_of_mr_labelling_errors}
\end{table*}

We also provided the following additional annotations:
\paragraph{MR order:} The order of the mentions of MR values in corresponding sentences
\paragraph{Number of sentences:} The number of sentences included in the text part
\paragraph{Sentence indexes:} An index of the sentences that include the corresponding MR values

Finally, the dataset consists of 40,560 examples for training, 4,489 for validation, and 4,555 for test.
We named it the {\it E2E refined dataset}.
An example from it is shown in \autoref{tab:example_of_the_e2e_refined_dataset}.

\begin{table*}[ht]
    \centering
    \begin{small}
    \begin{tabular}{l|p{12cm}}
    Refinement type&Text examples\\
    \hline \hline
    Irregular MR values & The {\bf Cotto} ranges between twenty and twenty-five pounds with a high customer rating. {\bf The Portland Arms} serves Italian food near the riverside.\\
    \hline
    Overlap & The Golden Curry served English food, {\bf is adult only}, is in the city centre, {\bf is adult only}, has a customer rating of 5 out of 5 and is near the Cafe Rouge.\\
    \hline
    Indefinite article & For {\bf an} child friendly, average coffee shop serving fast food try The Eagle, riverside near Burger King.\\
    \hline
    Typos & A kid friendly Strada has moderate price range and customer rating of 1 {\bf of of} 5.\\
    \hline
    British English & Alimentum providing fast food less than \pounds20 price range. It is located in city {\bf center}.\\
    \hline
    Currency mark & Close to Yippee Noodle Bar in the city centre a French restaurant, Alimentum, has low ratings but the price is less than {\bf 20lb}.\\   
    \hline
    Symbol & Browns Cambridge sells Indian food, and is kids friendly{\bf ,,} it is in riverside area near The Sorrento\\
    \end{tabular}
    \end{small}
    \caption{Examples of text refinement}
    \label{tab:examples_of_text_refinement}
\end{table*}

\section{Text Refinement}\label{sec:text_refinement}
\subsection{Error Correction}\label{subsec:error_correction}
We corrected the following types of errors in the original E2E dataset.
Some examples are shown in \autoref{tab:examples_of_text_refinement}.
Note that although our corrections may not be exhaustive, they will support future studies.

\paragraph{Irregular MR Values:}
Each attribute of an MR should have only one value or be empty.
However, some MR data have two values for the  \texttt{name} and \texttt{near} attributes.
We removed such irrelevant data from the dataset.

\paragraph{Overlaps:}
We removed a phrase that appears twice in a sentence.

\paragraph{Indefinite Articles:}
We corrected usage mistakes in the indefinite articles of ``a'' and ``an.'' 

\paragraph{Typos:}
We corrected over 3,700 typos in the MR values and sentences.

\subsection{Normalization}\label{subsec:normalization}
\paragraph{British English:}
Since the E2E dataset is based on British English, we replaced such words with American spellings as ``center'', ``flavor'', ``organize'', and ``traveling'' with ``centre'', ``flavour'', ``organise'', and ``travelling.''

\paragraph{Prices:}
\texttt{priceRange} is categorized into ``cheap'', ``moderate'', ``expensive'', ``lower than \pounds20'', ``\pounds20-25'', and ``more than \pounds30'' (\autoref{tab:all_variation_of_mr_values_in_the_e2e_refined_dataset}).
In that case, ``\pounds23'' should be labelled as ``\pounds20-25''.
However, the value of such prices as ``\pounds23'', ``\pounds22'', ``\pounds24'', and ``from \pounds20 to \pounds25'' in the sentences cannot be determined uniquely from the label ``\pounds20-25''.
To avoid this issue, we replaced all of the price values that should be labelled as ``\pounds20-25'' with ``\pounds20-25''.
The same idea was also applied to ``lower than \pounds20'' and ``more than \pounds30'' labels.

\paragraph{Currency Expressions:}
To simplify the expressions of currency units, we use ``\pounds20'' instead of ``20 pounds'', ``20gbp'', ``20lb'', ``20 quid'' and so on.

\paragraph{Symbols:}
We normalized such symbols as periods, commas, white spaces, etc.

\paragraph{Quotation Marks:}
We use single quotation marks instead of double quotations in the refined dataset.

\paragraph{Capital Letters:}
We fixed the capitalization errors in proper nouns and words that begin sentences.

\begin{table*}[t]
    \centering
    \begin{small}
    \begin{tabular}{l|r|p{10cm}}
    \hline
    Attribute & \# Variations & MR values (delexicalized) \\
    \hline
    \hline
    \texttt{Name} & 1 & NAME \\
    \hline
    \texttt{eatType} & 4 & (empty), coffee shop, pub, restaurant\\
    \hline
    \multirow{2}{*}{\texttt{food}}&\multirow{2}{*}{11}& (empty), American, Canadian, Chinese, English, fast food, French, Indian, Italian, Japanese, Thai\\
    \hline
    \texttt{priceRange}&7& (empty), \pounds20-25, cheap, expensive, less than \pounds20, moderate, more than \pounds30\\
    \hline
    \texttt{customer rating}&7& (empty), 1 out of 5, 3 out of 5, 5 out of 5, average, high, low\\
    \hline
    \texttt{area} & 3 & (empty), city centre, riverside\\
    \hline
    \texttt{familyFriendly} & 3 & (empty), no, yes\\
    \hline
    \texttt{near} & 2 & (empty), NEAR\\
    \hline
    \end{tabular}
    \end{small}
    \caption{All variations of MR values in the E2E refined dataset}
    \label{tab:all_variation_of_mr_values_in_the_e2e_refined_dataset}
    \vspace{6mm}
    \centering
    \small
    \begin{tabular}{l|l|l|c|c}
    \hline
    \multirow{11}{*}{MR}&Attribute&Value&Order&Sentence index\\
    \cline{2-5}
    &\multirow{2}{*}{\texttt{name}}&NAME&\multirow{2}{*}{1}&\multirow{2}{*}{1}\\
    &&(THE WRESTLERS)&&\\
    \cline{2-5}
    &\texttt{eatType}&restaurant&3&1\\
    \cline{2-5}
    &\texttt{food}&Italian&4&1\\
    \cline{2-5}
    &\texttt{priceRange}&moderate&2&1\\
    \cline{2-5}
    &\texttt{customer rating}&(empty)&0&0\\
    \cline{2-5}
    &\texttt{area}&city centre&5&2\\
    \cline{2-5}
    &\texttt{familyFriendly}&yes&7&3\\
    \cline{2-5}
    &\multirow{2}{*}{\texttt{near}}&NEAR&\multirow{2}{*}{6}&\multirow{2}{*}{2}\\
    &&(RAJA INDIAN CUISINE)&&\\
    \hline
    Number of sentences&\multicolumn{4}{l}{3}\\
    \hline
    \multirow{2}{*}{Text}&\multicolumn{4}{p{12cm}}{THE WRESTLERS is a moderately priced restaurant that serves Italian food. It is located in the city centre near RAJA INDIAN CUISINE. Great place to bring your family.}\\
    \hline
    \multirow{2}{*}{Text (delexicalized)}&\multicolumn{4}{p{12cm}}{NAME is a moderately priced restaurant that serves Italian food. It is located in the city centre near NEAR. Great place to bring your family.}\\
    \hline
    \end{tabular}
    \caption{Example of the E2E refined dataset: Original sample of the E2E dataset is shown in Table \ref{tab:example_of_the_e2e_dataset}.}
    \label{tab:example_of_the_e2e_refined_dataset}
\end{table*}

\section{MR Refinement}\label{sec:mr_refinement}
\subsection{Label Names}\label{subsec:label_names}
We replaced ``high'' with ``expensive'' for the  \texttt{priceRange} attribute.
For the \texttt{food} attribute, we defined additional labels: ``American'', ``Canadian'', ``Indian'', and ``Thai''.
The refined labels are listed in \autoref{tab:all_variation_of_mr_values_in_the_e2e_refined_dataset}.

\subsection{Labelling Errors}\label{subsec:labelling_errors}
We manually corrected the MR labelling errors throughout the E2E dataset.

\subsection{MR Order}\label{subsec:mr_order}
We annotated the order of the mentions of the MR values in the corresponding sentences in \autoref{tab:example_of_the_e2e_refined_dataset}.
If the MR value is empty, the order is represented by ``0''.

\subsection{Number of Sentences}\label{subsec:number_of_sentences}
We put the number of sentences in the text part, as shown in \autoref{tab:example_of_the_e2e_refined_dataset}.
We simply found the number of sentences using periods (``.'') and question marks (``?'').
Since the example in \autoref{tab:example_of_the_e2e_refined_dataset} shows that the text part includes three periods, we set its number of sentences to ``3''.

\subsection{Sentence Indexes}\label{subsec:sentence_indexes}
We also annotated each MR value with its appearance in the text by sentence-level indexes (\autoref{tab:example_of_the_e2e_refined_dataset}).
The example in \autoref{tab:example_of_the_e2e_refined_dataset} shows that the values of \texttt{eatType}, \texttt{area}, and \texttt{familyFriendly} appear in the first, second, and third sentences, respectively.
The index is set to ``0'' when an MR value is empty.

\section{Others}\label{sec:others}
\paragraph{Delexicalization:}
Because all of the \texttt{name} and \texttt{near} values appear as-is in the sentences, we replaced such values in the text and MR values with ``NAME'' and ``NEAR'' to standardize the data.
The original values are stored to keep the original information, although the standardized forms are still useful for training MR-to-text models.

\paragraph{Deduplication:}
We conducted deduplication of the MR-text pairs and excluded about 1,500 pairs from the dataset.

\section{Limitations}\label{sec:limitations}
Although we modified the E2E dataset for the development of MR-to-text models, the following limitations remain:
\begin{itemize}
    \item As mentioned in Section \ref{subsec:error_correction}, we removed the data that had irregular MR values. However, multiple values may be allowed under different formulations of MR-to-text problems for more complex situations.
    \item We currently ignored referring expressions, although generally they should be allowed.
    \item We regard attributes other than \texttt{name} as modifiers of a \texttt{name}. However, an attribute sometimes modifies \texttt{near}. Our current formulation ignores such relationships.
\end{itemize}

\section{Conclusion}\label{sec:conclusion}
We described our E2E refined dataset.
We reduced the deletion/insertion/substitution errors in the original E2E dataset and refined it by correcting errors and normalizing some expressions to simplify the sentences.
We also refined the annotation of the MR values and annotated the MR order, the number of sentences, and the sentence indexes as additional information.
The dataset and the data conversion programs in Python are publicly available at \url{https://github.com/KSKTYM/E2E-refined-dataset}.
We believe that this dataset will support future studies in related research fields.

\bibliography{anthology,custom}
\bibliographystyle{acl_natbib}

\end{document}